\documentclass{article}

\usepackage{aloha}
\usepackage{times}
\usepackage{pgfplots}
\usepackage{booktabs}
\usepackage{siunitx}
\pgfplotsset{compat=1.18}
\usepackage{subcaption}

\usepackage[utf8]{inputenc} % allow utf-8 input
\usepackage[T1]{fontenc}    % use 8-bit T1 fonts
\usepackage{url}            % simple URL typesetting
\usepackage{booktabs}       % professional-quality tables
\usepackage{amsfonts}       % blackboard math symbols
\usepackage{nicefrac}       % compact symbols for 1/2, etc.
\usepackage{microtype}      % microtypography
\usepackage{xcolor}         % colors

\usepackage{graphicx}
\usepackage{amsmath}
\usepackage{amssymb}
\usepackage{booktabs}
\usepackage{subcaption}
\usepackage{multicol}
\usepackage{multirow}
\usepackage{wrapfig}
\usepackage{pgf-pie}
\usepackage{enumitem}
\usepackage{placeins}
\usepackage{pgfplots}
\usepgfplotslibrary{statistics}
\usepackage{float}
\usepackage{siunitx}
\usepackage[table]{xcolor}
\usepackage{booktabs}
\usepackage{tikz}
\usepackage{pgf-pie}
\usepackage{xcolor}
\usepackage[colorlinks=true, linkcolor=blue, urlcolor=blue, citecolor=blue]{hyperref}
\usepackage{url}

\title{ShowUI-Aloha: Human-Taught GUI Agent}

\author{
Yichun Zhang \quad
Xiangwu Guo \quad
Yauhong Goh \quad
Jessica Hu \quad
Zhiheng Chen \quad
Xin Wang \quad
Difei Gao \\
Mike Zheng Shou\thanks{Corresponding author}\\
\vspace{0.6em}
\normalsize Show Lab, National University of Singapore\\
\url{https://showlab.github.io/Aloha_Page/}\\
}

\begin{document}

\maketitle

\begin{figure}[H]
\centering
\includegraphics[width=\textwidth]{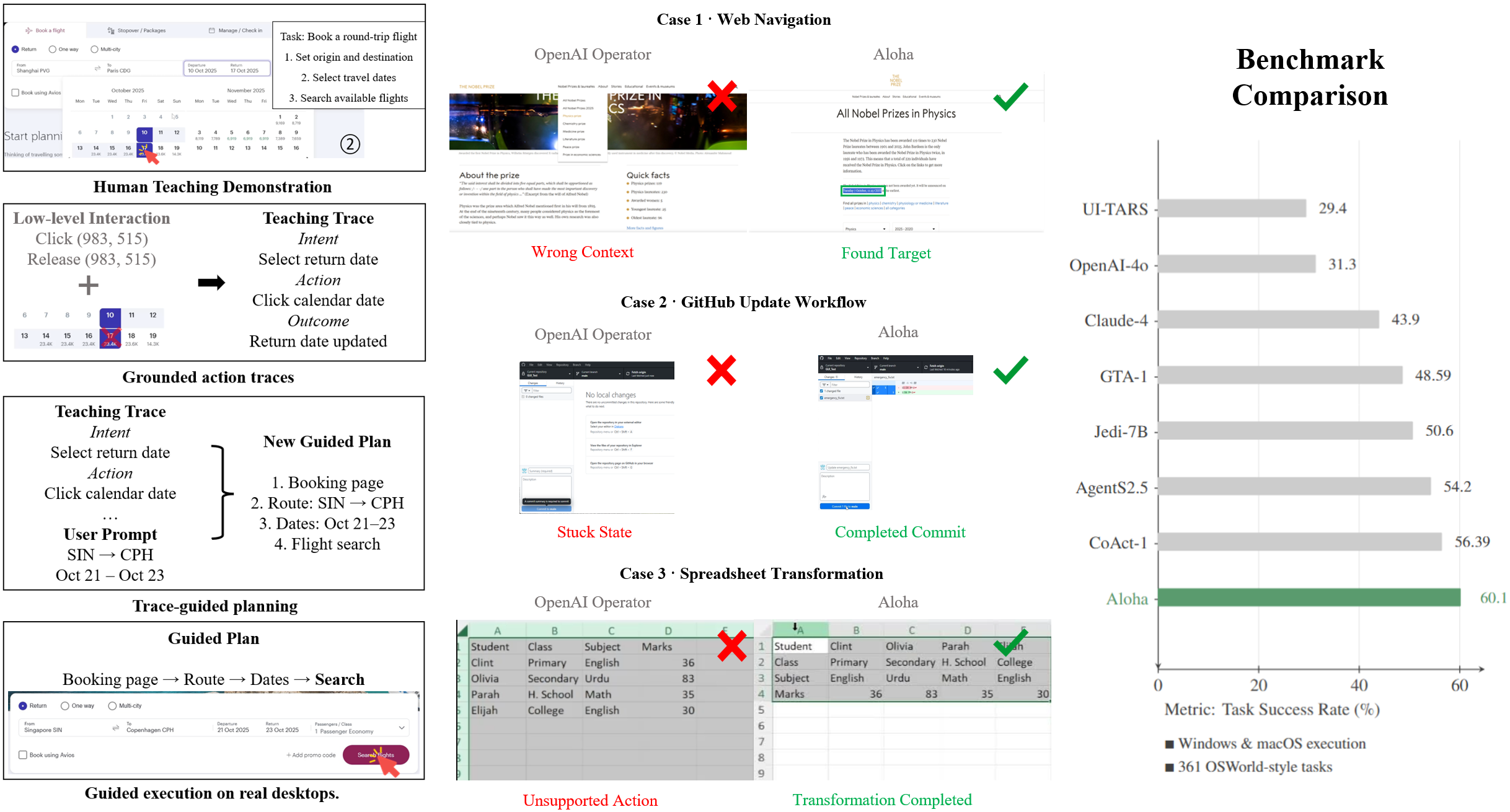}
\caption{%
  \textbf{Overview and evaluation of ShowUI-Aloha.}
  \textbf{Left:} Human-taught demonstrations are converted into grounded action traces, which are lifted into trace- and prompt-guided plans and executed on real desktop environments.
  \textbf{Middle:} Qualitative comparisons across representative multi-step desktop tasks show that Aloha avoids common failure modes of unguided agents, such as context drift, unsupported actions, and stuck states.
  \textbf{Right:} Quantitative comparison on 361 OSWorld-style tasks executed on Windows and macOS demonstrates that human-guided planning enables higher end-to-end task success than existing autonomous and agentic baselines.
  }
\label{fig:teaser}
\end{figure}

\begin{abstract}
\textit{Graphical User Interfaces (GUIs) are central to human-computer interaction, yet automating complex GUI tasks remains a major challenge for autonomous agents, largely due to a lack of scalable, high-quality training data. While recordings of human demonstrations offer a rich data source, they are typically long, unstructured, and lack annotations, making them difficult for agents to learn from.To address this, we introduce ShowUI-Aloha, a comprehensive pipeline that transforms unstructured, in-the-wild human screen recordings from desktop environments into structured, actionable tasks. Our framework includes four key components: \textbf{A recorder} that captures screen video along with precise user interactions like mouse clicks, keystrokes, and scrolls. \textbf{A learner} that semantically interprets these raw interactions and the surrounding visual context, translating them into descriptive natural language captions. \textbf{A planner} that reads the parsed demonstrations, maintains task states, and dynamically formulates the next high-level action plan based on contextual reasoning. \textbf{An executor} that faithfully carries out these action plans at the OS level, performing precise clicks, drags, text inputs, and window operations with safety checks and real-time feedback. Together, these components provide a scalable solution for collecting and parsing real-world human data, demonstrating a viable path toward building general-purpose GUI agents that can learn effectively from simply observing humans.}

% \begin{figure}[ht]
% \centering
% \includegraphics[width=0.9\linewidth]{teaser.png}
% \caption{\emph{ShowUI-Aloha learns from a single human demonstration and generalizes the workflow to new environments.}.}
% \label{fig:pipeline}
% \end{figure}

% \begin{figure*}[t]
% \centering
% \begin{subfigure}[t]{0.24\textwidth}
%     \centering
%     \includegraphics[width=\textwidth]{t_1.png}
%     \caption*{\footnotesize Info Gathering}
% \end{subfigure}\hfill
% \begin{subfigure}[t]{0.24\textwidth}
%     \centering
%     \includegraphics[width=\textwidth]{t_2.png}
%     \caption*{\footnotesize GitHub Editing}
% \end{subfigure}\hfill
% \begin{subfigure}[t]{0.24\textwidth}
%     \centering
%     \includegraphics[width=\textwidth]{t_3.png}
%     \caption*{\footnotesize Spreadsheet Editing}
% \end{subfigure}\hfill
% \begin{subfigure}[t]{0.24\textwidth}
%     \centering
%     \includegraphics[width=\textwidth]{t_4.png}
%     \caption*{\footnotesize Flight Booking}
% \end{subfigure}

% \vspace{0.4em}
% \caption{\textbf{ShowUI-Aloha performing a diverse set of real-world desktop workflows.}
% From left to right: online information gathering, GitHub repository editing, spreadsheet manipulation, and end-to-end air-ticket booking.}
% \label{fig:teaser}
% \end{figure*}
% \FloatBarrier

\end{abstract}
    
\section{Introduction}
\label{sec:intro}

% Background: GUI application and importance
Graphical User Interfaces (GUIs) have become the primary medium for human-computer interaction, enabling users to navigate and operate a wide range of digital environments—from web browsers and mobile applications to desktop software. Automating GUI tasks through autonomous agents offers significant potential to boost productivity, broaden access to digital tools, and lay the groundwork for advanced AI systems capable of adapting to dynamic environments \citep{deng2023mind2web, xie2025osworld, rawles2023androidinthewild}. Recent progress in vision-language models (VLMs) \citep{llava, wang2024qwen2,yang2024ariauivisualgroundinggui,qin2025uitars} and agent frameworks \citep{wu2024atlas, xie2025osworld, Gao_2024_CVPR} has further propelled developments in GUI automation. While these methods exhibit strong capabilities in grounding UI elements, their limited understanding of the underlying software logic continues to hinder task completion, particularly in complex workflows.

% Related work
Early attempts~\citep{deng2023mind2web,zhou2023webarena,Gao_2024_CVPR,agashe2024agent,abuelsaad2024agenteautonomouswebnavigation,li2024appagentv2advancedagent,li2023sheetcopilotbringingsoftwareproductivity,lai2024autowebglmlargelanguagemodelbased,hu2024dawnguiagentpreliminary,zhang2023appagentmultimodalagentssmartphone} introduce agent frameworks that utilized large language models (LLMs) to decompose user tasks and generate corresponding action plans. These methods typically operated in a zero-shot setting, relying entirely on the general knowledge encoded in LLMs, often sourced from web-based materials such as tutorials. Subsequent works~\citep{hong2024cogagent,lin2024showui,qin2025uitars,xu2025aguvisunifiedpurevision,yang2024ariauivisualgroundinggui,cheng2024seeclickharnessingguigrounding,you2024ferretuigroundedmobileui,wu2024atlas} progressed toward unified GUI vision-language-action models. These models are trained not only on large collections of static screenshots but, crucially, on \textbf{human}-labeled interaction trajectories, allowing them to directly map visual inputs and task instructions to GUI actions. Compared to the earlier LLM-based agents, these multimodal models exhibit substantially improved task planning abilities, highlighting the importance of \textbf{human knowledge} in advancing GUI agent development.

% Data challenge in the community
Nevertheless, a persistent challenge lies in the scalable collection of GUI automation data. Most publicly available datasets~\citep{deng2023mind2web,chen2024guicourse, miniwobplus, aitz, rawles2023androidinthewild,wu2024guiactionnarratordid,chen2025guiworldvideobenchmarkdataset,lin2024videogui} require extensive manual annotation and are largely limited to websites and mobile applications, where metadata is more readily accessible. Consequently, the performance of current methods remains constrained in more complex desktop environments. At the same time, we observe that knowledge workers routinely perform tasks on computers, generating vast amounts of real-world interaction data. This observation raises a compelling question: \textit{What if computers could learn to use software and perform digital tasks simply by observing how humans naturally and routinely accomplish them?} Such human demonstration data are plentiful, rich in contextual information, and faithfully capture authentic user behavior, offering tremendous potential to accelerate the development of capable and generalizable GUI agents.

% Detailed Challenge For human demonstration
However, collecting and utilizing human demonstration videos introduces several unique challenges. First, the absence of standardized data collection tools hinders the scalability of data acquisition. Moreover, the collected data—typically in the form of interaction trajectories, are inherently unannotated, making it difficult for models to interpret and learn from them. These trajectories usually consist of raw visual inputs, screen recordings, and low-level interaction data such as pixel-level click positions. Yet, for effective learning, models must grasp not only the semantics of these interactions but also the underlying user intentions. Additionally, knowledge workers often perform numerous tasks over extended periods, resulting in untrimmed recordings with no explicit task boundaries. Models must therefore learn to distinguish between interrelated actions that constitute a coherent task and unrelated actions that belong to separate workflows.

% Our work
To address these challenges, we present ShowUI-Aloha, a human-taught desktop agent that learns directly from in-the-wild demonstrations rather than curated UI labels or synthetic trajectories. Our key insight is that natural user interactions already contain rich intent, temporal structure, and visual grounding—if one can reliably capture and interpret them. ShowUI-Aloha adopts a record–parse–learn paradigm: lightweight instrumentation logs raw keyboard–mouse events and screen activity during a user’s normal workflow, and an inference pipeline transforms this noisy signal into a compact, semantically meaningful teaching trajectory. This trajectory abstracts away low-level pixels while retaining intent, enabling the agent to understand what was done and why.

Built on top of this representation, ShowUI-Aloha employs a planning and execution mechanism that generalizes the demonstration to new tasks and unseen UI states. By leveraging natural language abstraction and robust grounding against the live desktop environment, the agent learns transferable procedures that remain effective even when layouts shift, dialogs appear unexpectedly, or the task details differs from the original demonstration. This demonstration-driven paradigm enables ShowUI-Aloha to construct essential task-specific knowledge base while maintaining strong flexibility, emphasizing abstraction over memorization. This demonstration-driven paradigm yields a flexible, practical alternative to prior rule-based or template-driven GUI agents, while still being lightweight enough for deployment.

% Contribution
This work makes the following contributions:

\begin{itemize}
    \item \textbf{A learning pipeline that derives structured teaching trajectories.}  
    We develop a record--parse--learn framework that transforms raw, in-the-wild human desktop interactions into semantically grounded teaching trajectories.

    \item \textbf{A lightweight execution system for robust generalization.}  
    Building on the learned trajectories, we design a planner--actor mechanism that executes tasks on live desktop environments with robustness to UI drift, layout changes, and unexpected system states.

    \item \textbf{Comprehensive evaluation on OSWorld-style tasks~\cite{xie2025osworld}.} 
    We evaluate ShowUI-Aloha on a large-scale suite of OSWorld-style tasks spanning diverse desktop applications, demonstrating strong generalization and real-world applicability under a user-oriented evaluation protocol.

    \item \textbf{A fully open-source desktop agent.}  
    We release the entire ShowUI-Aloha framework—including recorder, learner, planner, actor, and evaluation tools—as an open-source project to support transparent, reproducible, and extensible research on GUI agents.
\end{itemize}

Together, these contributions establish ShowUI-Aloha as a practical and scalable foundation for demonstration-driven computer-use intelligence.

\begin{figure}[H]
\centering
\includegraphics[width=0.9\linewidth]{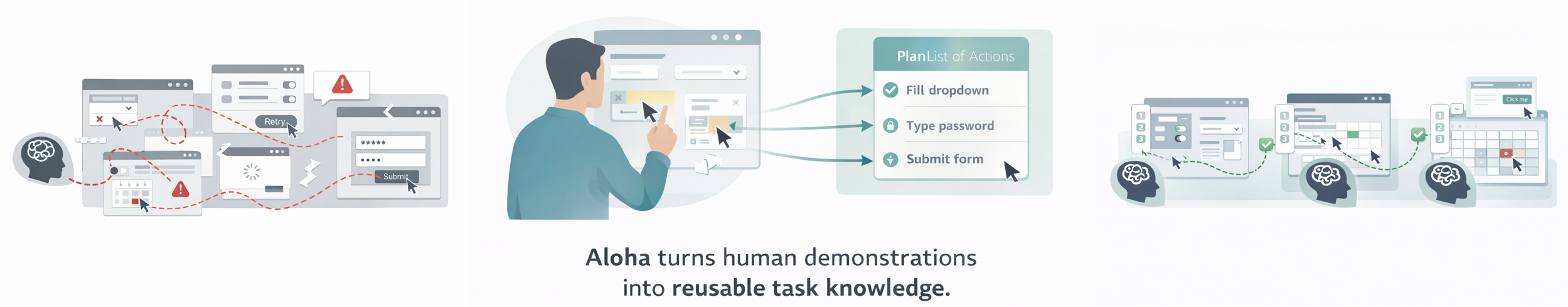}
\caption{
Overview of the Aloha paradigm for GUI agents.
Instead of relying on trial-and-error interaction, Aloha leverages a single human demonstration to distill reusable task guidance, which is then consistently applied to new task variants and interface layouts, enabling stable and generalizable execution across changing interfaces.
}
\label{fig:intro-4step}
\end{figure}

\section{Related Work}
\label{sec:relate_work}

\textbf{Computer Use Datasets.}
Recent progress of Large Language Models (LLMs) show promising potential beyond traditional text completion. Notably, agents~\cite{yao2023react,suris2023vipergpt,shen2023hugginggpt} showcase the capability to autonomously execute complex tasks through seamless tool integration.
This agentic capability has naturally extended to the digital GUI automation~\cite{gui_survey,gui_survey_2,gui_survey_3}. To power the development of GUI Agents, recent efforts have concentrated on the novel datasets and benchmarks across three representative platforms:
\textbf{(i) Website}~\cite{deng2023mind2web}, are often readily scalable due to the structured nature of HTML and available browser-based tools.  
However, the ease of automated collection can result in web corpora that are text-rich yet potentially noisy, characterized by a lack of rigorous human verification.
\textbf{(ii) Mobile}~\cite{aitz}, aims to enhance accessibility and interaction within simulated mobile environments like open-source Android and iOS.  Datasets in this domain, while valuable, tend to exhibit limitations in diversity, particularly in terms of software difficulties and action spaces.
\textbf{(iii) Desktop}~\cite{omniact,waa,xie2025osworld}, are considered highly valuable due to the inherent challenges in their collection. 
Unlike web and mobile platforms, desktop environments lack automated data collection pipelines.  The desktop interaction necessitates the complex integration of dense keyboard and mouse inputs. 
While benchmarks~\cite{waa,xie2025osworld,screenspotpro} offer evaluation frameworks for desktop platforms, scaling data collection to a large-scale remains a significant challenge. 
This underscores the critical need for the high-quality desktop GUI training corpora, especially those with human expert knowledge.
% Addressing this gap, to the best of our knowledge, our work introduces \textit{the largest, high-quality Desktop GUI dataset sourced from human demonstration videos} designed to support large-scale training.

\noindent\textbf{Learning from Human Demonstration.}
Learning from human demonstrations offers a data-efficient approach for training models in both physical~\cite{robomani,genh2r} and digital~\cite{juice,synatra} environments. 
Such demonstrations, often in the form of videos, serve as rich records of human experience and action for decision marking. 
Early works effectively leveraged demonstrations for action understanding from video~\cite{egovid} and for complex robot manipulation tasks~\cite{robomani,genh2r}, underscoring the paradigm's ability to learn complex behaviors without extensive hand-engineering. 
When it comes to the digital settings especially GUI, while many agent-training approaches rely on scaling data through automated crawling or synthetic generation \cite{adaptagent}, these methods can often lead to datasets lacking in quality and human-like strategic depth. 
Despite notable successes in adapting agents to new web or mobile environments~\cite{liu2025learnact,li2024appagentv2advancedagent} using demonstration learning and straightforward Supervised Fine-Tuning (SFT), we recognize that raw demonstration trajectories, being primarily records of low-level actions, frequently lack explicit encoding of high-level semantic user intentions or plans. 
To address this critical gap, our work introduces a carefully designed data workflow specifically to refine and augment these raw demonstrations, ultimately enabling more effective agent training by incorporating richer contextual understanding.
\section{Method}
\label{sec:method}

\begin{figure}[H]
\centering
\includegraphics[width=0.9\textwidth]{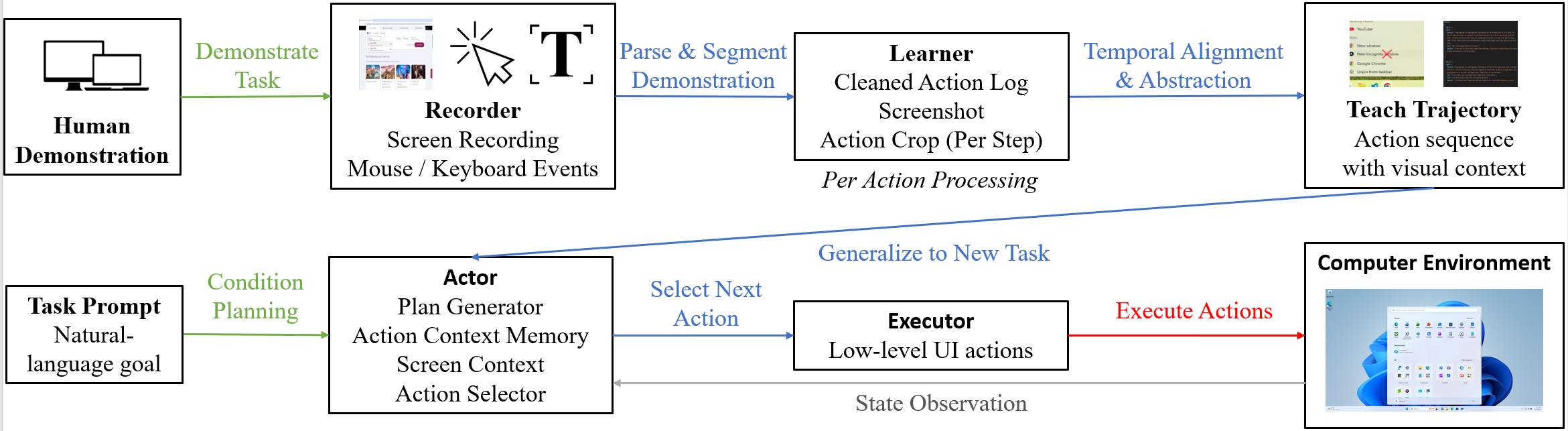}
\caption{\textbf{Overview of the Aloha workflow.} Human demonstrations are recorded and converted into structured action traces. The actor uses the task prompt and screenshots to generate an execution plan, while the executor performs each action on the computer. }
\label{fig:pipeline}
\end{figure}

Figure~\ref{fig:pipeline} shows an overview of the framework.
The goal is to unlock scalable data collection and parsing of human demonstrations, suitable for training and evaluating GUI models that can understand and potentially automate complex software workflows.
We introduce each core component: Recorder, Learner, Executor and Actor, in the following sections respectively.

% Sec.~\ref{sec:rec_tool} details the HumanDemoRecorder, a recording tool that collects videos and low-level action logs.
% Sec.~\ref{sec:parse_workflow} presents the HumanDemoParser, a parsing workflow that converts raw visual frames and low-level action logs into structured tasks and action sequence.
% After that, we introduce HumanDemoBench (Sec.~\ref{sec:data_eval}), which is curated from the data engine and will be used for assessing the HumanDemoParser and evaluating GUI models, respectively.

\subsection{Recorder}
\label{sec:rec_tool}
% \vspace{-10pt}
The recorder is implemented as an all-in-one portable software, which can be easily deployed in new Window or MacOS system with ease.
\begin{figure}[ht]
\centering
\includegraphics[width=0.9\linewidth]{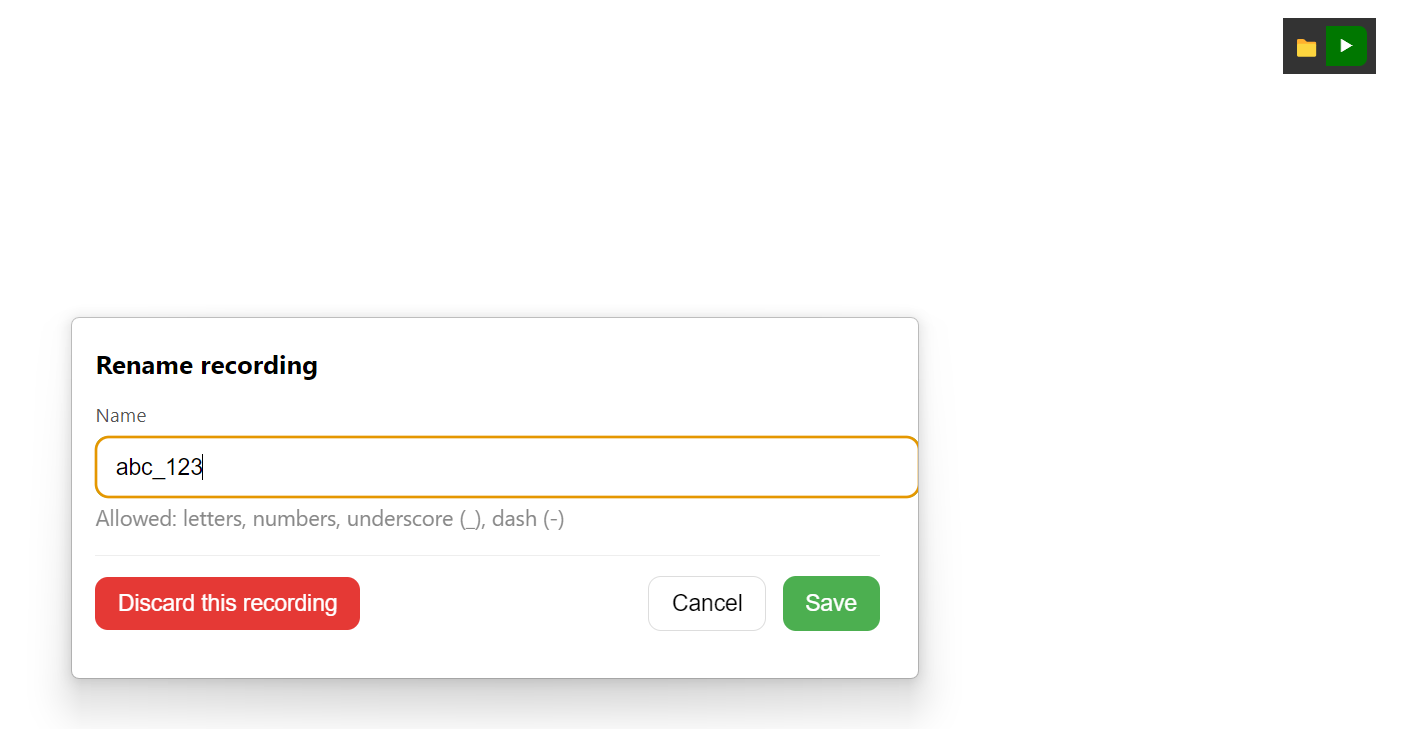}
\caption{User-facing interface of the ShowUI-Aloha Recorder.  
The recorder presents a minimal floating control panel (top right) for starting and stopping captures, while a modal dialog allows users to name or rename each recording with clear constraints on valid characters. These utilities support organized, large-scale data collection and facilitate downstream processing.
}
\label{fig:recorder}
\end{figure}

% \paragraph{Video Recording}
\noindent \textbf{Video Recording.}
The recorder aims to record videos with high frame rate, and simutaneously capturing detailed and dense user operations.
Considering the compatibility problem across machines, we employ FFmepg(avfoundation on MacOS as replacement) as the underlying video recording implementation, which is a widely-used, open-source multimedia framework.
When starting recording, the \texttt{ddagrab} filter in FFmpeg is triggered to capture and record the full-screen in full-resolution in 30 FPS (frame per second).

% \paragraph{Action Recording}
\noindent \textbf{Action Recording.}
During recording video frames, the tool simultaneously records the actions performed by user in real-time.
To implement this, we borrow code from KeyCastOW~\footnote{\url{https://github.com/brookhong/KeyCastOW}}, 
Instead of visualizing, our goal is to \textit{record} the action history.
Thus, in our implementation, the tool is designed to output the user actions into a logging file, together with the corresponding timestamps.
Specifically, it captures actions such as mouse clicks (left, right, wheel), mouse movements, drags, and keyboard inputs.
\begin{figure}[H]
\centering
\setlength{\tabcolsep}{3pt}
\begin{tabular}{ccc}
    \includegraphics[width=0.30\linewidth]{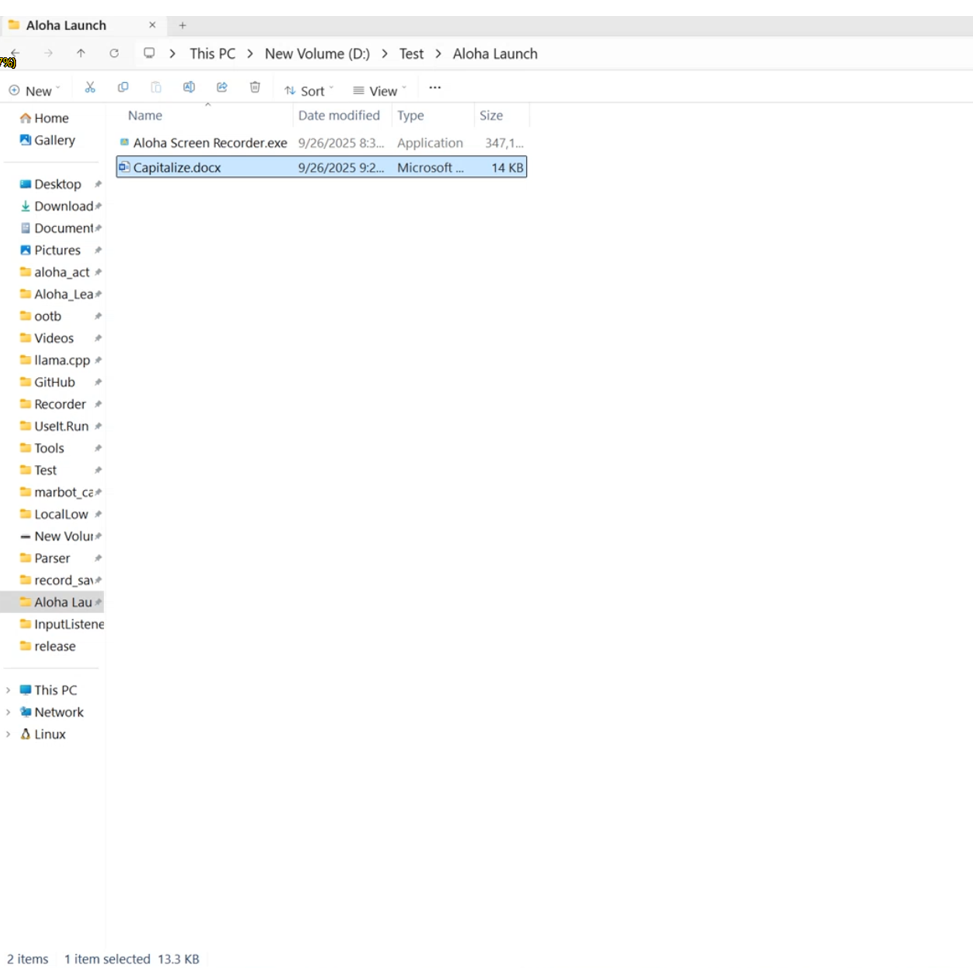} &
    \includegraphics[width=0.30\linewidth]{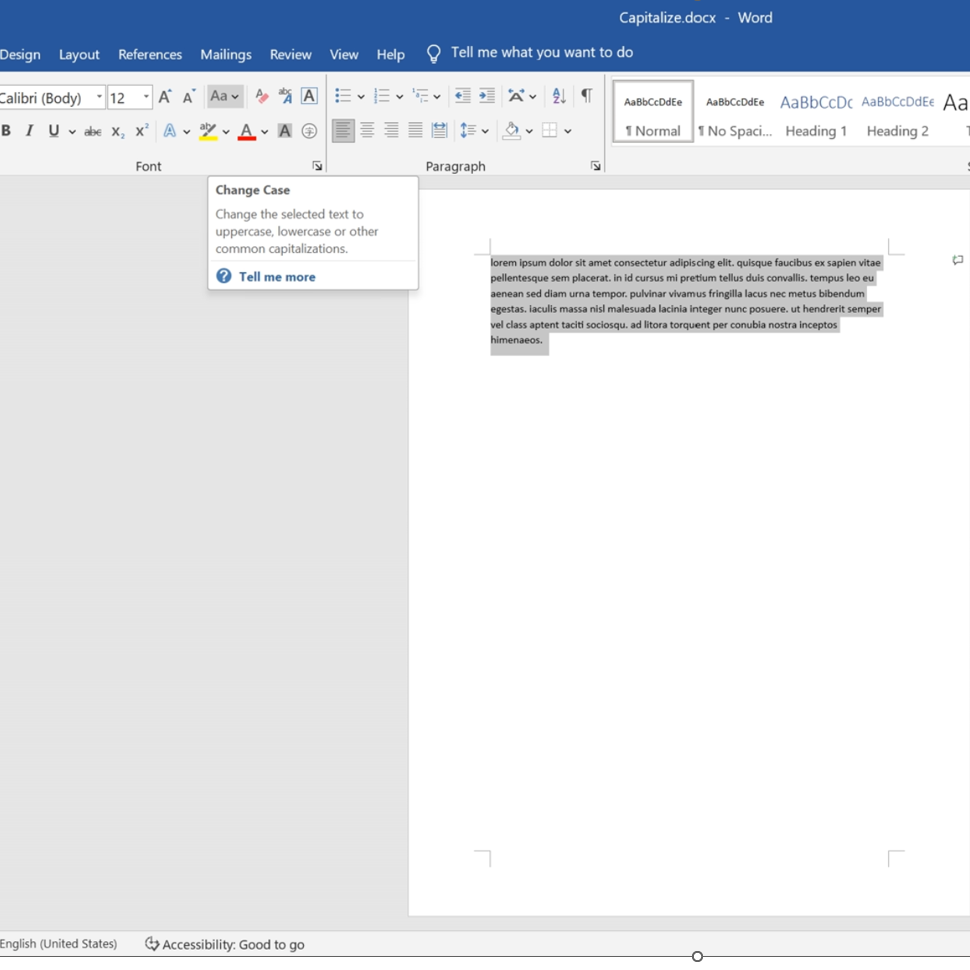} &
    \includegraphics[width=0.30\linewidth]{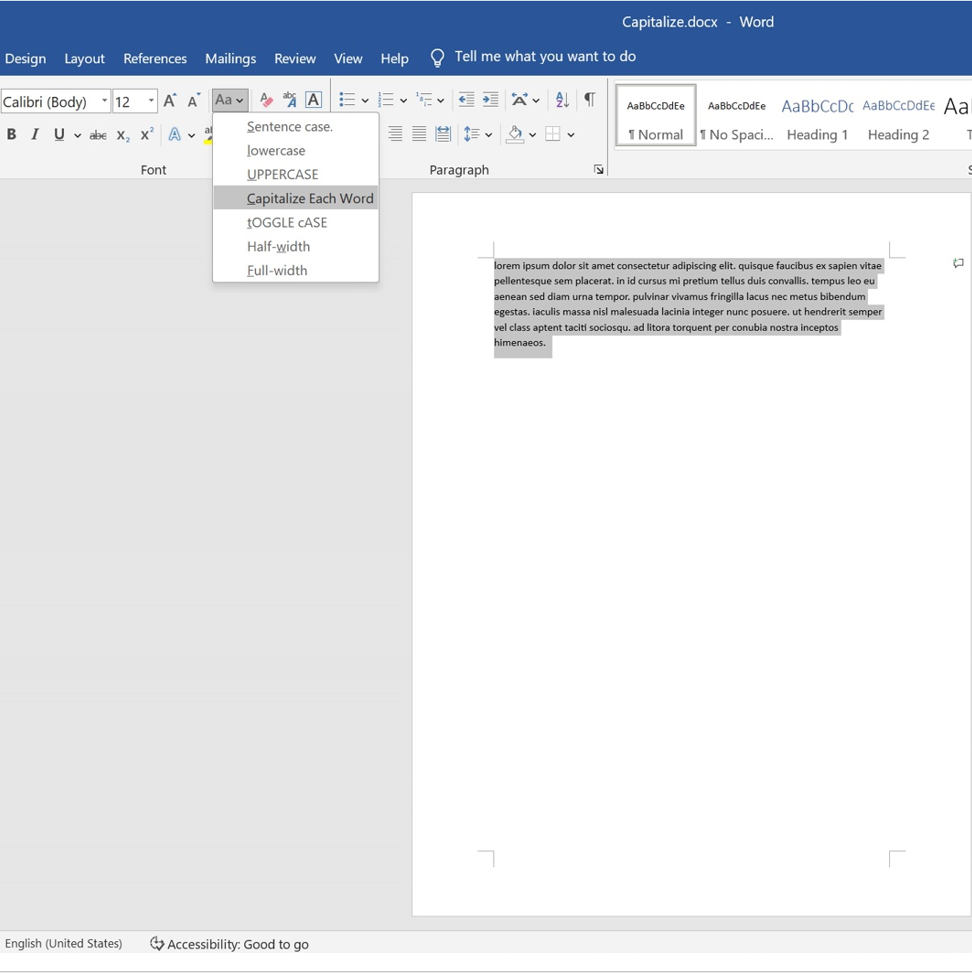} \\
    Raw frame - File Opening & Raw frame - Text Selection 2 & Raw frame - Format Selection
\end{tabular}
\caption{\textbf{Raw screen frames captured by the Aloha Recorder.}
These consecutive frames illustrate the natural visual trajectory present
in human desktop demonstrations. The Recorder captures full-resolution
frames at high frequency, preserving the fine-grained cursor dynamics,
UI transitions, and subtle motion patterns that are essential for downstream
action cleaning and trace generation.}
\label{fig:raw_frames}
\end{figure}
\begin{figure}[H]
\centering
\includegraphics[width=0.95\linewidth]{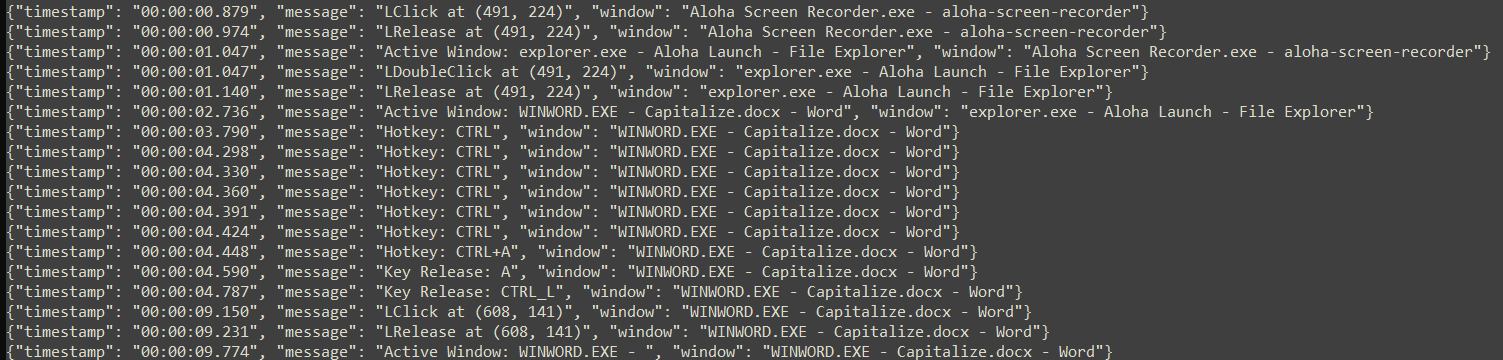}
\caption{\textbf{Raw action log captured by the Aloha Recorder of above frames.}
The recorder logs dense low-level input events such as mouse movements,
mouse down/up pairs, drags, and keystrokes with high-frequency timestamps.
This raw stream is noisy, redundant, and unstructured, reflecting natural
human behavior and motivating the need for the Action Cleaning stage in
the Aloha Learner.}
\label{fig:raw_log}
\end{figure}
% \paragraph{Integration}
\noindent \textbf{Integration.}
The complete recording pipeline—covering both screen capture and interaction logging—is packaged into a lightweight 170\,MB application with native builds for macOS and Windows. As shown in Fig.~\ref{fig:recorder}, the recorder includes a compact floating control panel for starting and stopping captures, together with a user-friendly renaming dialog that enforces consistent naming conventions for large-scale data collection. Additional utilities such as dynamic path configuration, one-click batch renaming, and API-call integration further streamline high-throughput experimentation and integration into broader systems. Beyond supplying high-quality teaching data for ShowUI-Aloha, this recorder also serves as a standalone tool for generating structured screen–action datasets for other GUI-related research tasks.

\FloatBarrier

\subsection{Aloha Learner}
\label{sec:learning_tool}
Aloha Learner converts raw human demonstrations into structured, semantic GUI action traces that can be executed and generalized by downstream modules. It consists of three tightly coupled components: a raw log parser, a screenshot marker, and a prompt-driven trace generator. Together, they bridge low-level human input with high-level, machine-actionable representations.

\noindent \textbf{Action Cleaning.}
The parser transforms the recorder’s high-frequency event stream into a compact sequence of semantic user actions. It first parses the raw log into primitive events—mouse down/up, motion, scroll, and keystrokes. Because the recorder samples at high temporal resolution, the initial stream contains redundant, fragmented, and noisy entries. The parser therefore applies a multi-stage consolidation pipeline. It merges consecutive keystrokes into coherent typing segments and reconstructs drag operations by linking press–move–release triples into continuous trajectories. Mouse down/up pairs are unified into single click events, with spurious single-clicks preceding double-clicks removed to avoid duplication. Scroll actions are normalized across input devices, and special keys such as Backspace and key combinations (e.g., Ctrl+S) are properly handled to faithfully reproduce the final text entered. All actions are then chronologically sorted and written to a cleaned log, yielding a minimal, semantically aligned sequence that accurately captures user intent.
\begin{figure}[H]
\centering
\includegraphics[width=0.9\textwidth]{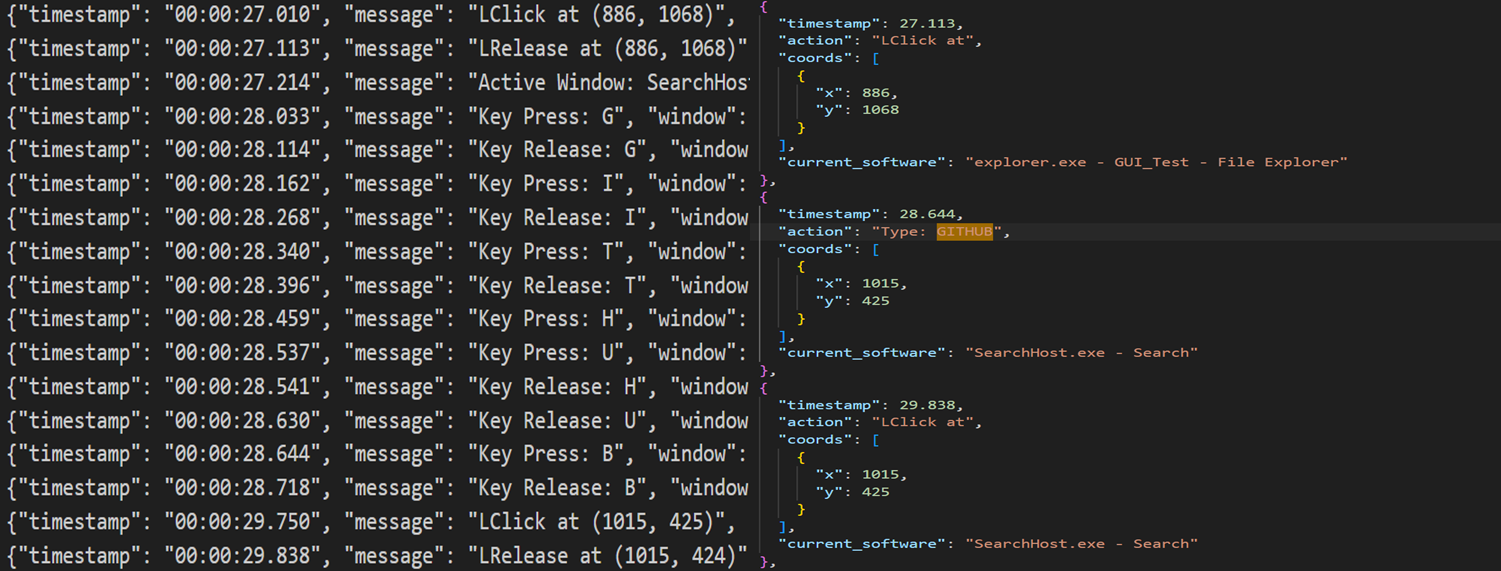}
\caption{\textbf{From raw events to grouped interaction primitives.}
The recorder logs dense, noisy, and highly redundant event streams (left).
Aloha Learner consolidates these signals into higher-level interaction primitives (right).}
\label{fig:raw_to_grouped}
\end{figure}

\begin{figure}[H]
\centering
\includegraphics[width=0.9\textwidth]{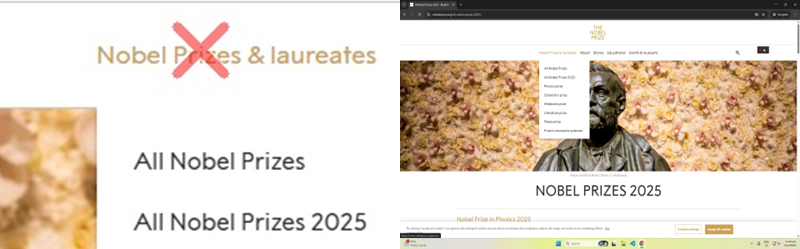}
\caption{Example of the zoomed-in and marked crop (left) and full-screen context (right) used by the trace generator.}
\label{fig:crop}
\end{figure}

\noindent \textbf{Marked Screenshot Generation.}
For each cleaned action, the screenshot marker produces synchronized visual inputs that make downstream reasoning more robust and unambiguous. Two images are extracted per action: a full-screen frame (typically 1920×1080) that provides global context, and a zoomed-in crop tightly centered around the interaction site. As shown by Figure~\ref{fig:crop}, to encode action semantics directly into the visual domain, expressive overlays are added: a semitransparent red 'X' for click-type events and a semitransparent red polyline indicating drag paths. These lightweight but informative markings remove the need for coordinate-level supervision and allow the system to attribute user intent directly from pixels. The size, semitransparency, and placement of these indicators are carefully calibrated to balance precise target localization with sufficient visual clarity, ensuring that downstream components can reliably analyze the underlying UI elements. This step forms a structured visual interface for the trace generator, ensuring that even ambiguous GUI regions become machine-interpretable. 

\noindent \textbf{Trace Generation.}
With the action list cleaned and screenshot pairs enriched with unobtrusive visual markers, the Trace Generator converts these multimodal signals into a coherent, stepwise natural-language trace. At its core, this module leverages a vision–language model (VLM) to jointly interpret the marked crop, the full-screen context, and the recent execution history, yielding semantically grounded descriptions that faithfully reflect user intent.
\begin{figure}[H]
\centering
\setlength{\tabcolsep}{6pt}
\begin{tabular}{cc}
    \includegraphics[width=0.42\linewidth]{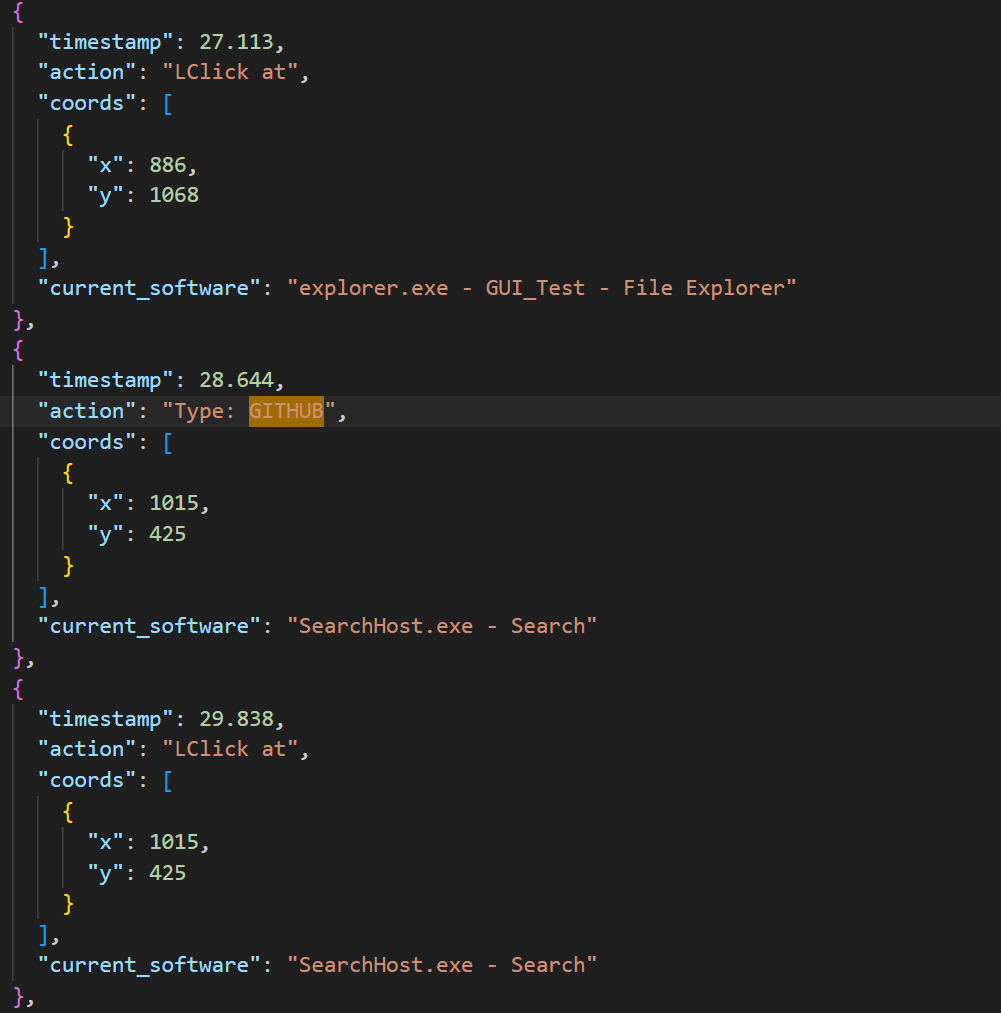} &
    \includegraphics[width=0.42\linewidth]{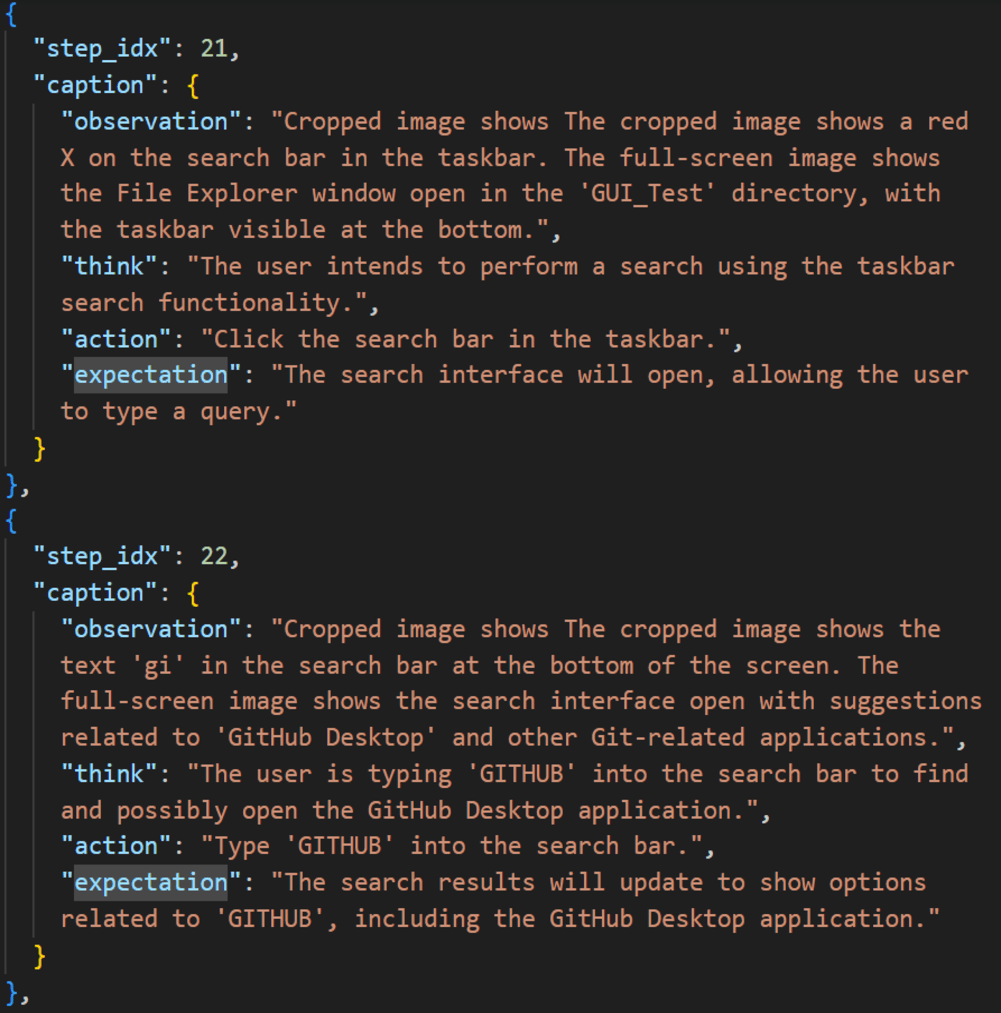} \\
    \small Grouped interaction primitives & \small Semantic, intent-aligned trace
\end{tabular}
\caption{\textbf{From grouped actions to semantic teaching traces.}
After low-level events are merged into coherent interaction primitives (left), 
Aloha Learner uses a vision--language model to reason over the marked screenshots, 
UI context, and recent action history to produce high-level semantic descriptions (right). 
Each step includes an \emph{Observation} of the UI state, a \emph{Think} field with 
brief reasoning, a normalized \emph{Action} such as ``click the File menu'' or 
``drag \texttt{pikachu.png} into dir2'', and an \emph{Expectation} describing how 
the interface should change. These semantic traces capture user intent and form 
the core supervision for downstream planning and execution.}
\label{fig:grouped_to_semantic}
\end{figure}

To initiate each step, the generator constructs a structured prompt consisting of: (1) a base instruction defining the expected JSON output schema; (2) an action-type–specific “delta” that injects concise priors about clicks, drags, scrolls, modifier keys, or typing; (3) a short summary of up to three previously generated steps; and (4) the high-resolution marked screenshots produced in the preceding stage. This combination allows the model to reason over both the localized interaction site—already visually annotated—and the broader UI state.

Upon receiving this prompt, the VLM produces a four-field caption containing an \emph{observation}, \emph{think} rationale, \emph{action} description, and \emph{expectation} of how the interface will change. A lightweight post-processing layer then sanitizes the output by removing spurious coordinate leakage, enforcing crop-first phrasing, and normalizing ambiguous or model-hallucinated operations. The resulting step is appended to the trajectory, and the process continues iteratively until all actions are consumed.

This design turns raw demonstrations into clean, executable traces without hand-crafted rules or task-specific templates. By grounding every step in both the marked visual context and preceding reasoning chain, the Trace Generator provides the Aloha Actor with a stable, interpretable, and semantically rich representation of human workflows.

The final output for each action is formatted as a four-field JSON record:
\textbf{Observation} (what is visually present),  
\textbf{Think} (brief reasoning and intent inference),  
\textbf{Action} (the concrete operation normalized by deltas), and  
\textbf{Expectation} (the immediate UI change that should occur).  
This structured representation provides both semantic clarity and operational determinism, enabling the downstream Aloha Actor to reliably execute, monitor, and recover from GUI interactions.

\noindent \textbf{Summary.}
Together, action cleaning, screenshot marking, and prompt-guided trace generation form the core of Aloha’s learner pipeline. They translate raw human–computer interaction into a coherent, machine-verifiable action plan—establishing the foundational vision–language interface through which Aloha acquires, understands, and reproduces real-world GUI tasks.

\subsection{Aloha Actor}
\label{sec:planning_tool}

\noindent \textbf{Aloha Actor} is the central orchestrator of the Aloha automation framework, coupling task-level reasoning with reliable GUI execution. It integrates the Aloha Planner as its cognitive front-end and coordinates multiple execution backends to operate robustly on real desktop environments. Together, they form a closed-loop system that plans, acts, verifies, and adapts—substantially beyond the capabilities of single-call or replay-based LLM agents.

\noindent \textbf{Aloha Planner.}
The planner interprets the user’s high-level goal, the instantaneous screenshot, and the demonstration-derived Guidance Trajectory. Its backbone is a large language model—commercial (e.g., GPT-4o, Claude), open-source, or locally deployed—which provides semantic priors but lacks intrinsic grounding of GUI states or demonstration structure. Aloha supplies this missing structure: each planning call is composed using structured templates that include screenshots, annotated action history, and step-aligned demonstration cues. The planner outputs a structured next-step plan with fields such as \textbf{Observation}, \textbf{Reasoning}, \textbf{Current Step}, \textbf{Action}, and \textbf{Expectation}, allowing trajectories to serve as soft references rather than rigid scripts. This enables consistent, goal-driven, and context-aware planning even under distribution shifts.

\noindent \textbf{Execution Backends as Primitive Actuators.}
Aloha is deliberately built \emph{above} existing foundation-model computer-use systems, not as a replacement for them. At the execution layer, the actor interfaces with general-purpose computer-use operators exposed by LLM platforms (e.g., OpenAI or Anthropic). These backends provide only low-level motor primitives—\emph{click}, \emph{double-click}, \emph{move}, \emph{scroll}, and \emph{type}—optionally accompanied by minimal perceptual grounding used solely to localize the target region on the screen. Critically, these operators do not perform task reasoning, trajectory following, goal verification, ambiguity resolution, or recovery. They execute one action at a time without maintaining any notion of progress or of the overarching task. In other words, they function strictly as interchangeable perception-assisted actuators.

\noindent \textbf{Actor Control Logic.}
Given the planner’s step plan, the Actor performs the integration and control logic required for reliable multi-step automation. It contextualizes the proposed action with environmental state—window hierarchy, OS behavior, application affordances—and chooses the appropriate backend invocation. When the external operator provides multiple possible click locations or ambiguous detections, the Actor resolves these cases using demonstration priors, UI topology, or deterministic heuristics. It selects fallback strategies such as hotkey execution when visual localization is unreliable, verifies post-action states, and triggers replanning when discrepancies arise. All visual and textual traces are logged and fed back into the reasoning loop, enabling stable and iterative correction over long horizons.

\subsection{Aloha Executor}
\label{sec:planning_tool}
% \vspace{-10pt}
\noindent \textbf{Aloha Executor} serves as the embodied control core of the Aloha framework, responsible for translating high-level agent intentions into precise, verifiable physical interactions on the screen. It forms the final link in the perception–planning–action chain, taking structured action commands from the Aloha Actor and turning them into actual mouse, keyboard, and window operations across heterogeneous desktop environments.

% \paragraph{Parsing}
\noindent \textbf{Parsing and Normalization.}
The executor first validates and parses the actor’s output through structured dispatchers. It supports a wide set of normalized action types—click, input, drag, scroll, key, hotkey, wait, and others—each mapped to specialized parser functions. These parsers handle schema variation and unify coordinates into absolute screen positions through a per-monitor offset system.

% \paragraph{Grounding}
\noindent \textbf{Coordinate Grounding and Safety.}
For multi-monitor setups, the executor computes per-screen offsets via platform-specific methods (screeninfo on Windows/Linux and Quartz on macOS). Every relative coordinate from the actor is converted into global screen coordinates before execution, ensuring consistent pointer behavior regardless of display configuration.

% \paragraph{Execution}
\noindent \textbf{Tool Execution and Feedback.}
Each parsed action is wrapped and dispatched asynchronously to the corresponding Computer Tool. This tool directly manipulates the desktop:
- Mouse Control: movement, click, double-click, drag, hover, and scrolling
- Keyboard Control: key sequences, hotkeys, and text typing
- Utility Actions: waiting, capturing screenshots, or reporting cursor position
Execution results are returned to the pipeline, providing structured runtime feedback to higher layers for logging and retry logic. In general, The Computer Tool encapsulates platform-independent GUI control, automatically handling coordinate scaling, multi-monitor bounding boxes, and interaction animation. It visually annotates clicks and drags through transient overlays for interpretability during demonstrations, and supports flexible scaling.

\section{Experiments}
\label{sec:experiment}
\FloatBarrier

Because ShowUI-Aloha operates in a human-taught setting, the official closed-loop OSWorld~\cite{xie2025osworld} evaluation pipeline cannot be applied directly.
We therefore evaluate Aloha by reinstantiating the complete OSWorld task suite using the released task specifications, manually executing \textbf{361 tasks} (excluding eight Google Drive–related tasks, as permitted by the OSWorld authors) with a \textbf{50-step budget} on real desktop environments.
\begin{figure}[t]
\centering
\includegraphics[width=\linewidth]{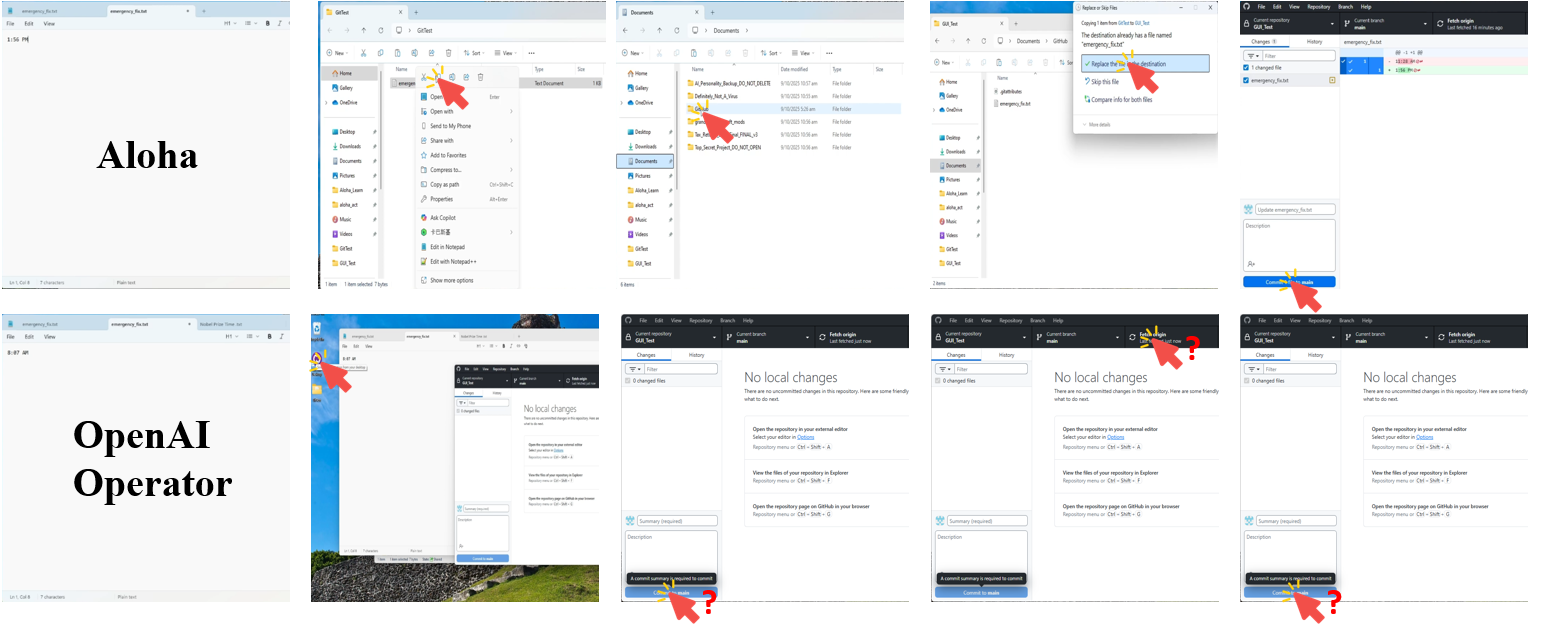}
\caption{\textbf{Qualitative comparison on a Git update workflow.}
Aloha follows the demonstrated procedure for propagating an edit from a scratch folder into a Git repository (top row): it modifies \texttt{emergency\_fix.txt}, navigates through \texttt{Documents} $\rightarrow$ \texttt{GitHub} $\rightarrow$ \texttt{GUI\_Test}, replaces the tracked file in the repository folder, and then issues a commit in GitHub Desktop.
In contrast, the unguided agent (bottom row) correctly edits and saves the local file, but then directly opens GitHub Desktop without first copying it into the repository path. Because the repo contains no changed files, it repeatedly tries to commit or push in an empty state and becomes stuck, illustrating a lack of procedural knowledge about intermediate file-management steps that Aloha inherits from human teaching traces.}
\label{fig:git_trajectory_comparison}
\end{figure}

\subsection{Experimental Setup}
\label{sec:exp_setup}

\noindent \textbf{Experiment Design.}Evaluations are conducted on both Windows and macOS platforms. Human testers operate in a controlled environment that replicates the original OSWorld setup. For each task, testers perform a new demonstration using a modified teaching prompt. For example, the original task “Copy all files matching *failed.ipynb in the current directory tree to ./fails while preserving the directory hierarchy” is adapted into “Copy the file ‘pikachu’ from my desktop photos folder to my desktop folder dir2.” This preserves the underlying task logic while altering filenames and paths. Such modifications are crucial to prevent the model from mechanically \textit{imitating} the teaching sequence, ensuring it instead generates a new action plan \textit{based on} the demonstration trace.

After each demonstration, the Aloha pipeline executes the task once, and success is measured using a binary score: 1 for successful completion and 0 otherwise. No partial credit is assigned. All 361 tasks are evaluated under this protocol, and the resulting data are collected and analyzed accordingly.

\noindent \textbf{Model and Actuator.}All experiments use \texttt{GPT-4o} as the vision–language model for perception and reasoning. 
Action execution is carried out through the OpenAI \texttt{Computer~Use} API, which serves as a lightweight, perception-assisted actuator. 
As detailed in the Methods section, this backend exposes only primitive OS-level operations (e.g., \emph{click}, \emph{move}, \emph{scroll}, \emph{type}) and executes them one at a time without maintaining task context. 
No additional reasoning or verification is performed by the actuator during evaluation.

\noindent \textbf{Test Devices.}
We evaluate ShowUI-Aloha on 3 representative desktop platforms to ensure cross-OS robustness. 
The first platform is a macBook Pro equipped with an Apple \texttt{M4~Pro} processor and \texttt{24\,GB} unified memory. 
The second is a Windows desktop with an Intel \texttt{i7–13700KF} CPU, \texttt{32\,GB} RAM, and an NVIDIA \texttt{RTX~4070\,Ti} GPU with \texttt{12\,GB} VRAM. 
The third is a Windows laptop (Lenovo Legion 5) equipped with an Intel \texttt{i7–10750H} CPU, \texttt{16,GB} RAM, and an NVIDIA \texttt{RTX~2060} GPU with \texttt{6,GB} VRAM.

This combination of macOS and Windows environments mirrors real-world GUI heterogeneity and provides a consistent evaluation base for OS-level manipulation tasks.

\subsection{Baseline Considerations and Comparison Setting}

\noindent \textbf{Lack of a Direct Baseline.}
Aloha operates in a \emph{demonstration-guided} regime, where the agent is explicitly taught the task procedure through a structured trajectory before execution. Unfortunately, no existing benchmark or prior work provides a comparable setting: current OSWorld baselines are all \emph{unguided, zero-shot} agents, and the official partial-credit scoring system inside the OSWorld evaluator cannot be reproduced outside the closed environment. As a result, it is not possible to construct a strict apples-to-apples baseline that matches Aloha’s supervision level, execution protocol, or evaluation metric.
\textit{We therefore report unguided agents only as contextual anchors rather than direct baselines.}

\noindent \textbf{Comparison to Unguided Agents.}
To contextualize Aloha’s capabilities, we report results from a representative set of \emph{unguided} computer-use agents spanning specialized GUI models, general-purpose foundation models, and multi-component agentic frameworks. These include vision–action models such as UI-TARS-1.5-7B~\cite{qin2025uitars} and OpenAI CUA 4o~\cite{openai2025o3}, generalist LLM baselines such as Claude 4 Sonnet~\cite{anthropic2025claude}, and recent agentic pipelines that coordinate reasoning and tool use, including GTA-1-7B w/ o3~\cite{yang2025gta1}, Jedi-7B w/ o3~\cite{xie2025scaling}, Agent S2.5 w/ o3~\cite{agashe2025s2}, and CoAct-1~\cite{song2025coact1}. These systems operate in a zero-shot, unguided setting, receiving only the task description without any human demonstration. In contrast, Aloha is explicitly provided the demonstration trajectory. Because these settings differ fundamentally in assumptions, supervision, and evaluation protocol, the reported numbers are \textbf{not directly comparable}. Instead, they serve to position Aloha within the broader landscape of GUI-agent paradigms.

\noindent \textbf{Evaluation Metric.}
The official OSWorld benchmark reports a continuous score in [0,1], where partial credit is granted using task-specific reward functions embedded in the closed evaluation environment. Since these reward functions are not publicly available, they cannot be reproduced outside the official runner. Consequently, we evaluate all systems—including our reproduction of unguided baselines—using a \textbf{strict binary success metric}: a task is counted as successful only if the final state exactly matches the goal, with no partial credit for intermediate progress. This yields a more conservative assessment of Aloha’s capabilities relative to OSWorld’s partially-credited scoring.

\subsection{Experimental Results}
\label{sec:results}
Table~\ref{fig:osworld_by_category} and Figure~\ref{fig:fig3} report Aloha’s performance across the ten OSWorld application categories. Aloha demonstrates strong generalization to diverse real-world software, achieving the highest success rates on \textit{Chrome} (91.3\%), \textit{OS operations} (83.3\%), and \textit{Thunderbird} (80.0\%). It also performs reliably on development and media applications, including \textit{VS Code} (73.9\%), \textit{LibreOffice Writer} (69.6\%), \textit{GIMP} (65.4\%), and \textit{VLC} (64.7\%). More complex office-style workflows such as \textit{Calc} (57.4\%) and \textit{Impress} (42.6\%) show moderate difficulty, while tasks requiring cross-application coordination remain the most challenging (\textit{Multi-apps}, 37.6\%).

Across all 361 evaluated tasks, Aloha successfully completes 217, corresponding to an overall success rate of \textbf{60.1\%}, demonstrating robust performance under diverse GUI environments and heterogeneous task structures.
\begin{figure}[H]
\centering
\fboxsep=12pt
\fboxrule=0pt  % set to 0.5pt if you want a visible card border
\fbox{%
\begin{minipage}{0.7\linewidth}

  % Title + subtitle
  {\large\textbf{Aloha Performance Across OSWorld Categories}}\\[2pt]
  {\footnotesize Strong coverage on everyday productivity workflows, with near-perfect success in browser tasks.}\\[10pt]

  % Table
  \renewcommand{\arraystretch}{1.25}
  \rowcolors{3}{gray!2}{white} % light zebra striping for readability

  \begin{tabular}{%
      >{\raggedright\arraybackslash}p{3.4cm}
      S[table-format=3.0]
      S[table-format=3.0]
      >{\raggedleft\arraybackslash}p{1.8cm}
    }
    \toprule
      \textbf{Category} & \textbf{Solved} & \textbf{Total} & \textbf{Rate} \\
    \midrule
      % High performers highlighted a bit more
      \rowcolor{green!8}
      Chrome               & 42 & 46 & 91.3\% \\
      \rowcolor{green!4}
      OS                   & 20 & 24 & 83.3\% \\
      \rowcolor{green!4}
      Thunderbird          & 12 & 15 & 80\% \\

      % Mid-high
      LibreOffice Writer   & 16 & 23 & 69.6\% \\
      GIMP                 & 17 & 26 & 65.4\% \\
      VLC                  & 11 & 17 & 64.7\% \\
      VS Code              & 17 & 23 & 73.9\% \\

      % Lower, more challenging categories
      \rowcolor{yellow!8}
      LibreOffice Calc     & 27 & 47 & 57.4\% \\
      \rowcolor{yellow!8}
      LibreOffice Impress  & 20 & 47 & 42.6\% \\
      \rowcolor{red!6}
      Multi-apps           & 35 & 93 & 37.6\% \\
    \midrule
      \rowcolor{blue!8}
      \textbf{Overall} & \textbf{217} & \textbf{361} & \multicolumn{1}{>{\raggedleft\arraybackslash}p{1.8cm}}{\textbf{60.1\%}} \\
    \bottomrule
  \end{tabular}

  \vspace{6pt}
  {\footnotesize\color{gray!70}
    OSWorld covers 361 real desktop tasks; Aloha currently automates 217 of them end-to-end.
  }

\end{minipage}
}
\caption{
ShowUI-Aloha demonstrates consistently stronger end-to-end task success than prior unguided and agentic GUI agents across a broad set of OSWorld-style real-world tasks.
}
\label{fig:osworld_by_category}
\end{figure}

\begin{figure}[H] \centering \begin{tikzpicture} \begin{axis}[ ybar, ymin=0,ymax=100, bar width=7pt, width=\linewidth, height=8cm, enlarge x limits=0.06, ymajorgrids, ylabel={Success Rate (\%)}, symbolic x coords={Chrome,GIMP,Calc,Impress,Writer,Multi-apps,OS,Thunderbird,VLC,VSCode}, xtick=data, x tick label style={rotate=45, anchor=east}, nodes near coords, nodes near coords align={vertical}, ] \addplot coordinates { (Chrome,91) (GIMP,65) (Calc,57) (Impress,43) (Writer,70) (Multi-apps,38) (OS,83) (Thunderbird,80) (VLC,65) (VSCode,74) }; \end{axis} \end{tikzpicture} \caption{Aloha OSWorld Task success rate in each category.} \label{fig:fig3} \end{figure}
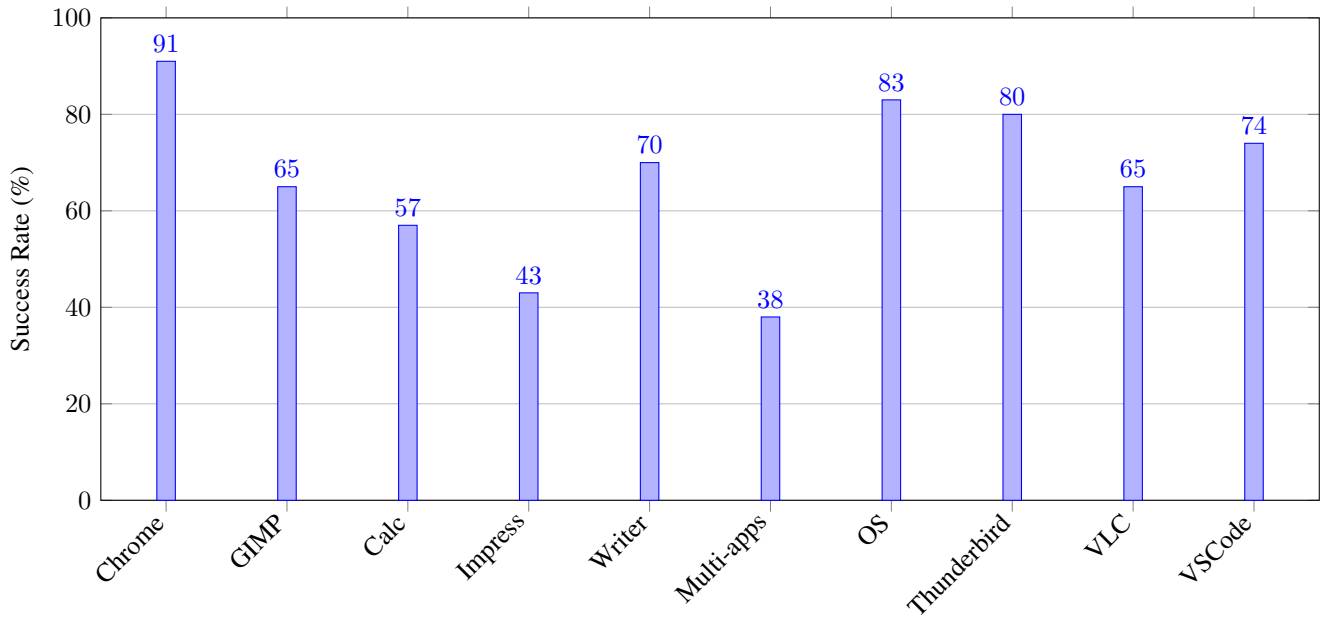

\begin{figure}[t]
\centering

\fboxsep=12pt
\fboxrule=0pt % set to 0.4pt if you want a light card border
\fbox{%
\begin{minipage}{0.95\linewidth}

% Title + subtitle
{\large\textbf{Teaching Mode Improves Performance over Strong Baseline Agents}}\\[2pt]
{\footnotesize Human-taught demonstrations enable ShowUI-Aloha to achieve higher end-to-end task success than unguided and agentic models on OSWorld-style tasks.}\\[10pt]

\renewcommand{\arraystretch}{1.25}
\setlength{\tabcolsep}{6pt}

\begin{tabular}{%
    >{\raggedright\arraybackslash}p{3.1cm}
    >{\raggedright\arraybackslash}p{2.4cm}
    S[table-format=2.2]
}
\toprule
\textbf{Model} & \textbf{Type} & {\textbf{Score}} \\
\midrule
UI-TARS-1.5-7B     & Specialized & 29.40 \\
OpenAI CUA 4o      & Specialized & 31.30 \\
Claude 4 Sonnet    & General     & 43.90 \\
GTA-1-7B           & Agentic     & 48.59 \\
Jedi-7B            & Agentic     & 50.60 \\
Agent S2.5         & Agentic     & 54.20 \\
CoAct-1            & Agentic     & 56.39 \\

\midrule
\rowcolor{green!12}
\textbf{ShowUI-Aloha} & \textbf{Human Guided} & \textbf{60.1} \\
\bottomrule
\end{tabular}

\vspace{6pt}
{\footnotesize\color{gray!70}
  OSWorld baseline models report graded benchmark scores; Aloha uses binary end-to-end success rate, making its performance strictly harder to achieve.
}

\end{minipage}
}

\caption{
\textbf{ShowUI-Aloha outperforms strong prior GUI agents, including fully autonomous and agentic systems, on OSWorld-style tasks.}
Human-taught task traces enable consistent improvements in end-to-end task success under a user-oriented evaluation protocol, highlighting a reliable and scalable pathway toward real-world desktop automation.
}
\label{tab:table2}
\end{figure}

\subsection{Error Analysis}
\label{sec:error}
Despite the overall strong performance, Aloha still exhibits several characteristic failure modes.
The most frequent error arises from failed or incorrect element selection, particularly in applications such as GIMP and LibreOffice Impress where multiple visually similar icons exist in dense toolbars. Because the underlying model receives only limited textual or structural descriptions of each icon, it occasionally confuses semantically related actions. Enhancing visual-text alignment and incorporating richer metadata could mitigate this ambiguity.
Many failures arise from the agent’s imprecise drag-selection when editing text. Without semantic awareness of text boundaries, the motor-level controller often selects too little, too much, or nothing at all before deleting or typing. These slight inaccuracies lead to leftover characters, malformed inputs, and incomplete value replacement, revealing the sensitivity of text manipulation to precise cursor-drag behavior.
These failures show that while Aloha reliably executes most deterministic GUI workflows, its remaining weaknesses center on fine-grained element localization and precise drag-based text editing. Addressing these challenges will require stronger icon-level semantic grounding and more adaptive recovery mechanisms for subtle selection and editing drift.

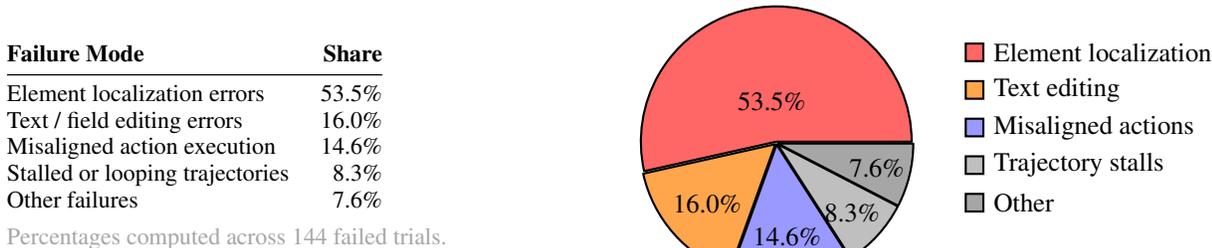
\begin{figure}[H]
\centering

\fboxsep=12pt
\fboxrule=0pt
\fbox{%
\begin{minipage}{0.95\linewidth}

{\large\textbf{Breakdown of Failure Modes in Unsuccessful Trials}}\\[2pt]
{\footnotesize Element localization dominates the error distribution, revealing the clearest opportunity for targeted improvement.}\\[10pt]

\begin{minipage}{0.40\linewidth}
\footnotesize
\begin{tabular}{@{}l r@{}}
\textbf{Failure Mode} & \textbf{Share} \\
\midrule
Element localization errors      & 53.5\% \\
Text / field editing errors      & 16.0\% \\
Misaligned action execution      & 14.6\% \\
Stalled or looping trajectories  & 8.3\% \\
Other failures                   & 7.6\% \\
\end{tabular}

\vspace{4pt}
{\footnotesize\color{gray!70} Percentages computed across 144 failed trials.}
\end{minipage}
\hfill
\begin{minipage}{0.55\linewidth}
\centering

\begin{tikzpicture}
\pie[
    radius=1.8,                 % compact pie chart
    explode=0.02,               % light separation
    color={red!60,orange!70,blue!40,gray!50,gray!70},
    text=inside,                % Percentage inside slices
    text=legend
]{
    53.5/Element localization,
    16.0/Text editing,
    14.6/Misaligned actions,
    8.3/Trajectory stalls,
    7.6/Other
}
\end{tikzpicture}

\end{minipage}

\end{minipage}
}

\caption{Breakdown of Failure Modes in Unsuccessful Trials}
\label{fig:failure_pie}
\end{figure}

\subsection{Ablation Study}
\label{sec:ablation}

To assess the contribution of core components in Aloha, we conduct an ablation study on a representative subset of 30 OSWorld tasks evenly covering all ten application categories. Each variant uses the same evaluation protocol and environment as the main experiments. We focus on two major factors: (1) the role of human demonstration traces (\textit{TeachTrace}), and (2) the role of the planner’s temporal memory (\textit{PlannerMemory}), which conditions each action on both the current step and the sequence of previous decisions. For the planner ablation, we disable all contextual functions, reducing the agent to a one-step decision-maker with no accumulated task history.

\begin{figure}[H]
\centering

\fboxsep=12pt
\fboxrule=0pt  % set to 0.4pt if you want a visible card border
\fbox{%
\begin{minipage}{0.95\linewidth}

% Title + subtitle
{\large\textbf{Impact of Human Teaching and Planner Memory}}\\[2pt]
{\footnotesize Ablation on a 30-task OSWorld subset shows both components are critical for stable long-horizon execution.}\\[10pt]

\renewcommand{\arraystretch}{1.25}
\setlength{\tabcolsep}{8pt}

\begin{tabular}{%
    >{\raggedright\arraybackslash}p{3.2cm}
    S[table-format=2.1]
    S[table-format=1.2]
}
\toprule
\textbf{Variant} & {\textbf{Success Rate (\%)}} & {\textbf{Step-Norm}} \\
\midrule
\rowcolor{green!12}
\textbf{Full (ours)}      & \textbf{63.3} & \textbf{0.89} \\
-- TeachTrace             & 36.7          & 0.56          \\
-- PlannerMemory          & 50.0          & 0.68          \\
\bottomrule
\end{tabular}

\vspace{6pt}
{\footnotesize\color{gray!70}
  ``Success Rate'' is exact task completion; ``Step-Norm'' is mean normalized progress (ReachedStep / TraceStep) across 30 OSWorld tasks.
}

\end{minipage}
}

\caption{Impact of Human Teaching and Planner Memory.}
\label{tab:ablation}
\end{figure}

Removing the human \textit{TeachTrace} produces the largest degradation, dropping success from 63.3\% to 36.7\% and reducing normalized progress from 0.89 to 0.56. This confirms that demonstration-driven procedural grounding is the primary contributor to Aloha’s reliability on multi-step OSWorld tasks. Disabling \textit{PlannerMemory} also yields a substantial decline (50.0\% success, 0.68 Step-Norm), indicating that temporally aware planning is essential for maintaining task-state consistency and avoiding drift in longer sequences. Taken together, the ablations show that Aloha’s improvements arise from the combination of human-taught trajectories and a memory-equipped planner—neither component alone is sufficient for robust, generalizable GUI automation.
\begin{figure}[H]
\centering
\includegraphics[width=0.92\linewidth]{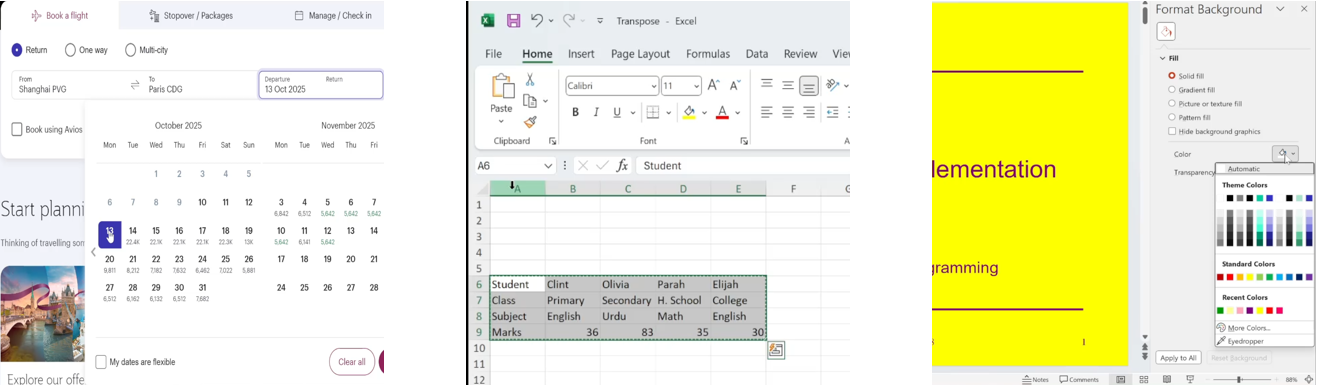}
\caption{\textbf{Additional examples of Aloha executing complex real-world tasks.}
From left to right: (1) automated air-ticket booking involving multi-step UI navigation
and structured form filling; (2) advanced Excel operations such as matrix transposition
and cell-range manipulation; and (3) batch editing of slide backgrounds in PowerPoint.
These diverse tasks demonstrate Aloha's ability to generalize beyond simple click-and-type
patterns and reliably follow high-level workflows across heterogeneous applications.}
\label{fig:thumbnail_multitask}
\end{figure}

% \begin{figure}[t]
% \centering
% \includegraphics[width=0.7\linewidth]{ablation_examples.png}
% \caption{
% \textbf{Qualitative ablation results.}
% Left: successful execution with the full model.
% Middle: failure without TeachTrace---the agent misinterprets procedural intent.
% Right: failure without ScreenInfo---the agent clicks a visually similar but incorrect icon.
% These cases highlight the complementary roles of human demonstrations and visual grounding.
% }
% \label{fig:ablation_examples}
% \end{figure}

\FloatBarrier
\section{Conclusion}

This work presents \textbf{ShowUI-Aloha}, a practical framework that converts raw human desktop demonstrations into structured, executable trajectories for GUI agents. Aloha integrates a lightweight cross-platform recorder, a learner that distills interaction logs into intent-aligned traces, and a planner–actor system that uses temporal memory to operate low-level OS actuators—enabling the agent to learn via \emph{abstraction rather than memorization}.

Evaluated on a broad set of OSWorld-style tasks, Aloha demonstrates robust multi-step execution across diverse real-world desktop applications. A single demonstration often generalizes to an entire \emph{task group} sharing the same workflow logic, and ablations confirm the importance of both human-derived traces and temporally conditioned planning.

\textbf{Limitations.} Remaining challenges include fine-grained icon disambiguation, noise-sensitive drag-based text selection, and reliance on at least one demonstration per workflow family.

\textbf{Future Work.} Expanding task-group coverage, improving icon-level and text-structure understanding, and scaling toward few-shot or demonstration-free generalization are promising directions. Turning Aloha’s structured traces into compact vision–language–action models may further reduce runtime dependence on demonstrations.

Because \textbf{ShowUI-Aloha is fully open-sourced}, researchers can readily swap VLMs, actuators, or planners, or reuse individual components such as the recorder. We hope Aloha serves as a foundation for developing demonstration-driven, memory-aware GUI agents capable of reliable operation in complex desktop environments.

% \section*{Acknowledgements}
% We thank Mingyu Ouyang and Siyuan Hu for their contributions to the early internal version of the algorithm, which informed parts of this open-source implementation.

\newpage
{
    \small
    \bibliographystyle{plain}
    \bibliography{main}
}

\end{document}